# NiyamAI - Cryptographically Verifiable Guardrails for Autonomous LLM Agents

Aditya Katkar
Om Karkele
Kartik Mandhane
Manisha More
Yash Kashid
*Vishwakarma Institute of Technology, Pune, Maharashtra, India*

## Abstract

Autonomous LLM agents with tool execution capabilities introduce severe security risks through prompt injection, goal hijacking, and unauthorized action invocation. Existing guardrails rely on unverified, host local software filters — system prompts, semantic classifiers, policy engines — that share the execution environment of the untrusted agent, offering no guarantee to an external observer that a safety policy was correctly evaluated. A compromised host produces no evidence of its own failure.

This paper presents NiyamAI, an intent bound runtime guardrail architecture providing cryptographically verifiable execution integrity for autonomous agents. At session initialization, permitted tools and operational constraints are sealed into an immutable Intent Contract under a SHA256 commitment. Every tool invocation is intercepted by a deterministic authority gate and classified by a dedicated neural Judge (11→8→2 feedforward network). For each authorized action, NiyamAI generates a succinct zkSNARK proof certifying correct policy evaluation under the committed contract; execution proceeds only after that proof verifies.

Across 2,000 AgentSafetyBench scenarios under 5fold stratified crossvalidation with out of fold scoring, NiyamAI achieves 88.8% F1 at a 1.0% false positive rate (bootstrap 95% CI [85.5%, 92.1%]), against 66.8% for Llama Prompt Guard 2, 46.2% for GPTOSSSafeguard, and 40.4% for NeMo Guardrails; McNemar's exact test confirms each margin at $p < 0.0001$. Proof generation adds 1.7 s per approved action, verification 51 ms, with an 18.6 KB proof — verifiable by any third party without access to model parameters. We further subject NiyamAI's own enforcement mechanism to 18 adversarial vectors across six classes, disclosing two implementation vulnerabilities identified and remediated during development.

## 1. INTRODUCTION

Large Language Models are no longer confined to conversational tasks. Agentic systems now interact with external environments, invoke tools, execute commands, query databases, and take actions that directly modify real-world state. This expansion in capability is matched by an expansion in attack surface: prompt injection, unauthorized tool invocation, and hallucination driven actions become materially consequential once an agent can alter system state or reach sensitive resources.

Contemporary agent safety rests on trust. A system prompt constrains the model's behavior. An output filter inspects its responses. A policy engine mediates between the agent and its tools. Each of these mechanisms assumes the platform executing them is honest and unmodified — an assumption that fails under a compromised host, a misconfigured deployment, or an adversary who has circumvented the filter itself. In that failure mode there is no residue: the safety check simply did not run, and nothing in the system records that it should have.

This paper pursues a stronger guarantee. Rather than asserting that enforcement occurred, NiyamAI produces evidence that it occurred. Each session begins by binding the agent's permitted tools and operational constraints into a cryptographically sealed Intent Contract. Every subsequent tool invocation is intercepted, evaluated against that contract by an isolated Judge model, and — if authorized — accompanied by a zero-knowledge proof certifying that the evaluation was performed and produced the recorded decision. The tool executes only after the proof verifies. Blocked actions are recorded in a tamper evident, hash chained ledger.

The design choice that makes this computationally practical is scope. Proving inference over a multibillion parameter LLM remains infeasible; proving inference over a compact Judge model — 11 inputs, 8 hidden units, 431 arithmetic constraints — completes in approximately 1.7 seconds on commodity hardware, with verification in under 51 ms. Verification requires neither the model's parameters nor trust in the host that executed it.

The primary contributions of this paper are as follows:

1. **Intent Binding Protocol**. A SHA256 commitment scheme that seals an agent's permitted tools and operational constraints at session initialization, preventing midsession capability drift. Post-seal modification produces an immediate hash mismatch on reverification.
2. **Intent Bound Interception Layer**. Tool requests are intercepted before execution and evaluated

against the sealed contract through a deterministic authority gate, a payload inspector, a session-bound control flow guard, and a neural Judge model.

3. **Verifiable ZK-ML Execution**. Each authorized decision yields a Halo2KZG zkSNARK proof generated via EZKL. Execution is gated on successful verification, replacing hostlevel trust with an artifact any third party can check independently.

Empirical evaluation and adversarial analysis. On 2,000 AgentSafetyBench scenarios under leakage free 5fold cross-validation, NiyamAI attains 88.8% F1 at a 1.0% false-positive rate, outperforming NeMo Guardrails, Llama Prompt Guard 2, and GPTOSSSafeguard at $p < 0.0001$ under McNemar's exact test. We additionally attack NiyamAI's own enforcement mechanism across 18 vectors in six classes, disclosing two implementation vulnerabilities found and remediated during development.

## 2. LITERATURE REVIEW

The development of NiyamAI was influenced by three areas of research, autonomous AI agents, AI safety mechanisms, and ZK-ML.

### 2.1 Autonomous Agents and LLM Security Vulnerabilities

LLMs have transcended mere generation of text to perform as autonomous agents that plan tasks and interact with external systems. Wang et al. [14] provided a survey on the architecture of agents based on LLMs and their capability to work in dynamic settings. Schick et al. [13] developed Toolformer, which proved the possibility of language models learning to make use of external APIs during task execution.

This autonomy brings about novel threats to security. Liu et al. [15] analyzed the prompt injection attack and demonstrated the possibility of users' inputs affecting the behavior of AI agents. Prompt injection and insecure plugins become the important security issues for LLM applications [18]. According to Nasr et al. [16], adaptive jailbreak attacks can circumvent a large volume of software regardless of guardrails.

Evaluation of agent safety has been supported by purposebuilt benchmarks. AgentSafetyBench [25] provides 2,000 scenarios spanning eight risk categories — data leakage, property loss, physical harm, and others — each with declared tool environments, enabling systematic measurement of an agent's behavior under adversarial and ambiguous instructions. We use it as our evaluation substrate in Section IV.

### *2.2 AI Alignment and Governance*

Bai et al. [17] introduced Constitutional AI to encourage models to follow predefined principles during their responses. In parallel, frameworks such as the NIST AI Risk Management Framework [19] and ISO/IEC 42001 [20] have been developed to help organizations manage the risks associated with AI deployment.

While these efforts meaningfully advance alignment and governance, they fall short of providing strong runtime guarantees. Enforcement tends to rely on model behavior or server-side controls, neither of which is immune to modification or bypass, since no cryptographic verification is in place. As Nasr et al. [16] point out, adaptive attacks can take advantage of exactly these gaps, finding ways around software-based safeguards that lack deeper structural protections.

A parallel line of work provides runtime guardrails for deployed LLM systems. NVIDIA's NeMo Guardrails [24] implements programmable rails that route user input through LLM based self-check prompts before permitting downstream action. Meta's Llama Prompt Guard 2 and OpenAI's GPTOSSSafeguard take a classifier-based approach, applying a purpose trained safety model to incoming requests. These systems are effective at detecting recognizably harmful content, and we evaluate against all three in Section IV. They share the limitation this paper addresses: each produces a decision, not evidence of a decision, and each executes within the same trust boundary as the agent it constrains.

### 2.3 Verifiable Computation and ZK-ML

Cryptographic verification has emerged as an important approach for strengthening security guarantees in modern computing systems. The theoretical foundations of Zero-Knowledge Proofs (ZKPs) and interactive proof systems were established by Goldwasser et al. [6], [7] and refined into practical succinct noninteractive arguments (zkSNARKs) by Groth [8] and BenSasson et al. [23].

Recently, these cryptographic protocols have been applied to neural networks. In their survey of verifiable machine learning, Zhang et al. [4] showed how ZKPs can be used to verify inference while keeping model parameters private. Work by Lee et al. [2] and Fan et al. [3] showed that zkSNARKs can be integrated with smaller Convolutional Neural Networks (vCNNs) to support verifiable inference. Building upon this, South et al. [1] outlined generalized frameworks for the verifiable evaluation of machine learning models. The tooling for these implementations has significantly matured, notably with the EZKL system [11], which allows developers to compile standard ONNXformat neural networks directly into verifiable ZK circuits.

### 2.4 Relationship to Constrained Decoding Frameworks

A distinct line of work enforces output structure through token level logit manipulation during generation — constrained decoding frameworks such as Outlines and Guidance restrict the model's vocabulary at each decoding step to conform to a grammar or schema. These approaches operate INSIDE the language model's generation loop, at the token level, and are well suited to enforcing syntactic structure (e.g., valid JSON output). NiyamAI addresses an architecturally distinct problem: verifying the SEMANTIC INTENT of a fully formed tool call AFTER the LLM has decided what action to take, independent of how that decision was generated or which model produced it. This distinction matters practically: constrained decoding requires whitebox access to the generating model's logits at inference time, whereas NiyamAI's interception layer operates on any agent's output regardless of the underlying model or generation method, at the cost of not being able to prevent syntactically malformed generation in the first place. The two approaches are complementary rather than competing: a production system could reasonably use constrained decoding to ensure a tool call is well formed, and NiyamAI to verify that the well-formed call does not violate the session's declared intent.

### 2.5 The Gap in Current Research

Despite rapid advancements in ZK-ML [1], [2], [11], generating a cryptographic proof for a modern, multibillion parameter Large Language Model (LLM) remains computationally prohibitive for real time inference. Consequently, orchestration frameworks such as LangChain [12] mediate tool access through application level permission checks, which provide no external evidence that the check executed These soft check mechanisms are vulnerable to prompt injection, context window manipulation, and runtime tampering.

There is a critical gap between the theoretical promise of ZK-ML and the practical security needs of autonomous LLM agents. Our proposed architecture, NiyamAI, bridges this gap. By utilizing standard cryptographic commitment to agent intent (SHA256) [10] to seal an agent's intent, and applying ZK-ML solely to a lightweight, independent "Judge" model rather than the LLM itself, we introduce a computationally feasible framework for cryptographically enforcing and verifying agentic tool execution.

## 3. METHODOLOGY

NiyamAI is a framework for enforcing IntentBound Execution in autonomous AI agents. Its goal is to prevent critical actions from being executed without proof that they satisfy the defined safety policy. The architecture consists of three main components: Agent Layer, Guardrail Layer, and Verification Layer, as shown in Figure 1.

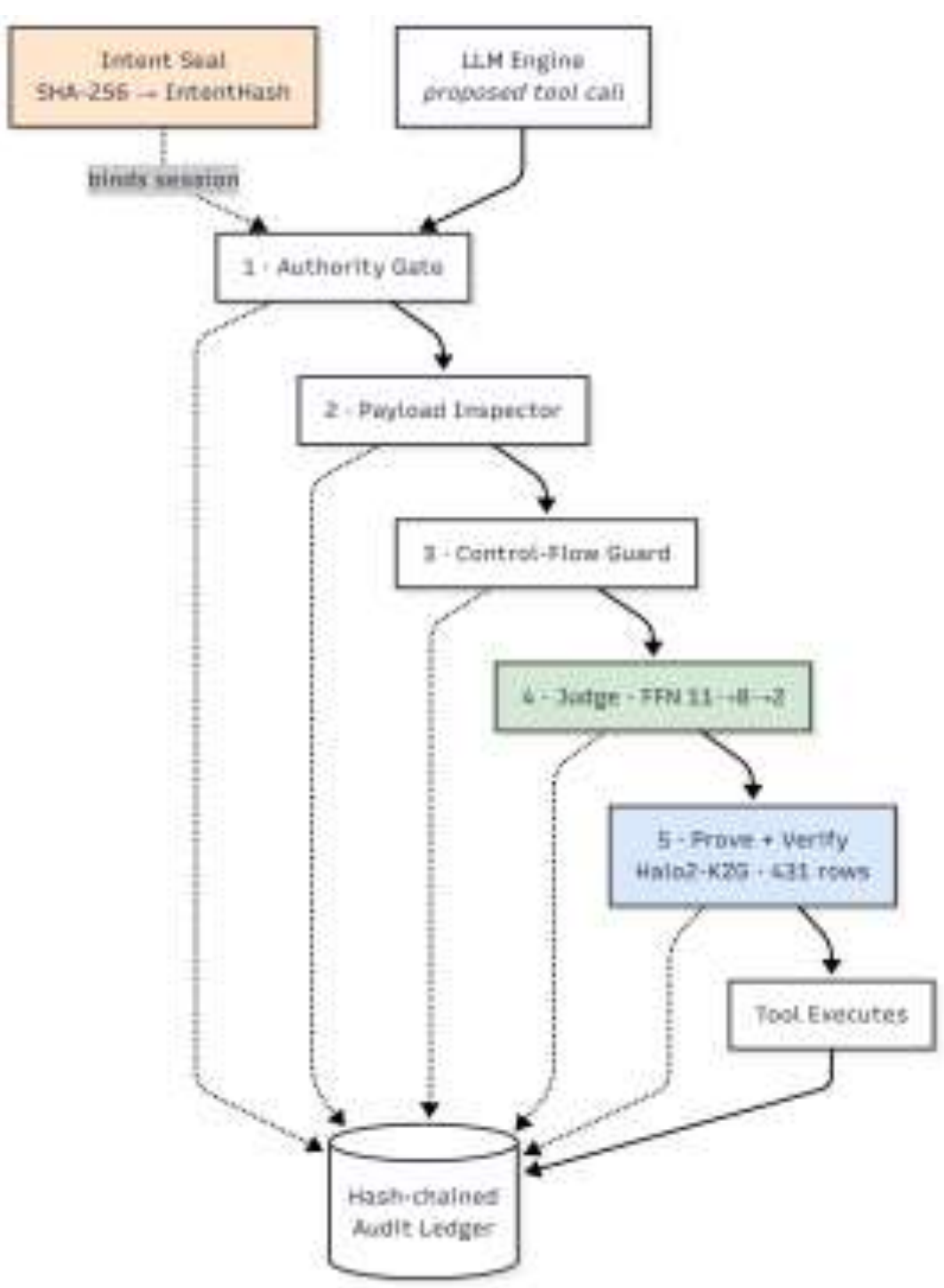

***Fig. 1.*** *High-level system architecture of Niyam-AI. Five enforcement layers precede execution; the tool runs only after proof verification..*

### 3.1 Intent Binding and Cryptographic Sealing

The basis for NiyamAI is the definition of the "Intent Contract" that is immutable and defined during initialization of the agent session. In contrast to the conventional prompt [15] which is vulnerable to context window tampering and prompt injection attacks, the Intent Contract is defined in a structured JSON format, listing the permissible tools, permissible data scope, and operational limitations.

After initialization, the Intent Contract is transformed to a standard format and then hashed using SHA256 [10]. The IntentHash will be constant throughout the session and will be used to verify all actions by the tools. Any alteration to the agent capabilities results in a hash mismatch, which in turn, blocks the action.

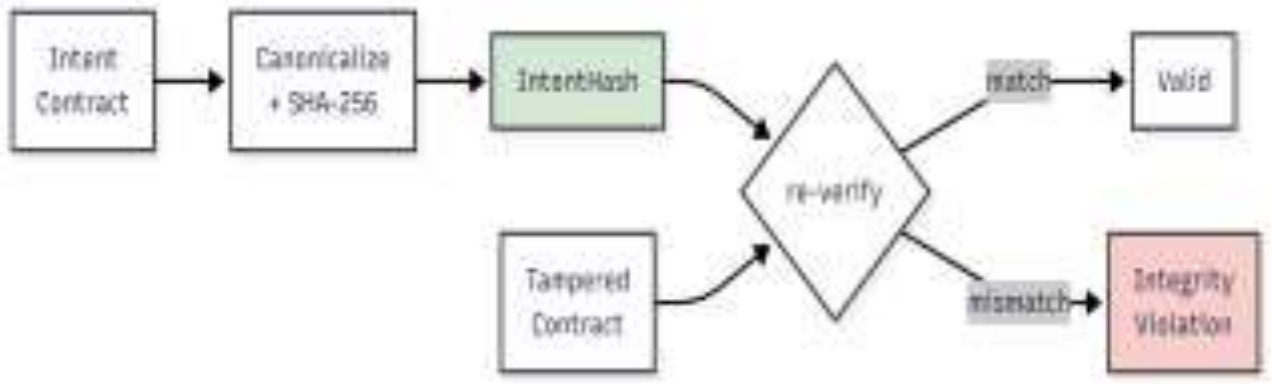

***Fig. 2.*** *The cryptographic sealing mechanism generating the immutable IntentHash..*

### 3.2 The Operation Layer (Local execution)

The operation layer is responsible for processing user inputs and generating proposed actions using localized, finetuning large language models. Unlike standard implementations that rely on high level agentic frameworks like LangChain framework [12] to manage the continuous loop of reasoning, planning, and tool selection. The core intelligence is powered by an open source Large Language Model (e.g., Llama3 or Mistral) [14].
The system implements a custom Reasoning Interception Loop. When a user provides a prompt, the local LLMs performs internal reasoning to determine the necessary operations. If the model determines that an external tool is required (following a structured format similar to Toolformer [13]), the system's execution engine intercepts the output string before it can be parsed or executed by the host operating system.

### *3.3 The Guardrail Layer (Interception and Classification)*

The Guardrail Layer acts as the security middleware. It utilizes a lightweight neural network-based classification engine that evaluates the ActionHash against the IntentHash. When a tool call is intercepted, the system extracts the tool's name and its proposed arguments to compute an ActionHash (via SHA256) that is cryptographic commitment to agent intent. The interception module then concatenates the session's IntentHash with the ActionHash to form an evaluation tuple. This tuple is fed into an isolated, lightweight Judge Neural Network (implemented in PyTorch) designed to classify the proposed action as Safe (1) or Unsafe (0).
If the Judge model classifies the action as "Unsafe," the execution is immediately blocked, and the intrusion attempt is recorded in an append only audit logger. If the action is classified as "Safe," the system proceeds to the proving phase.

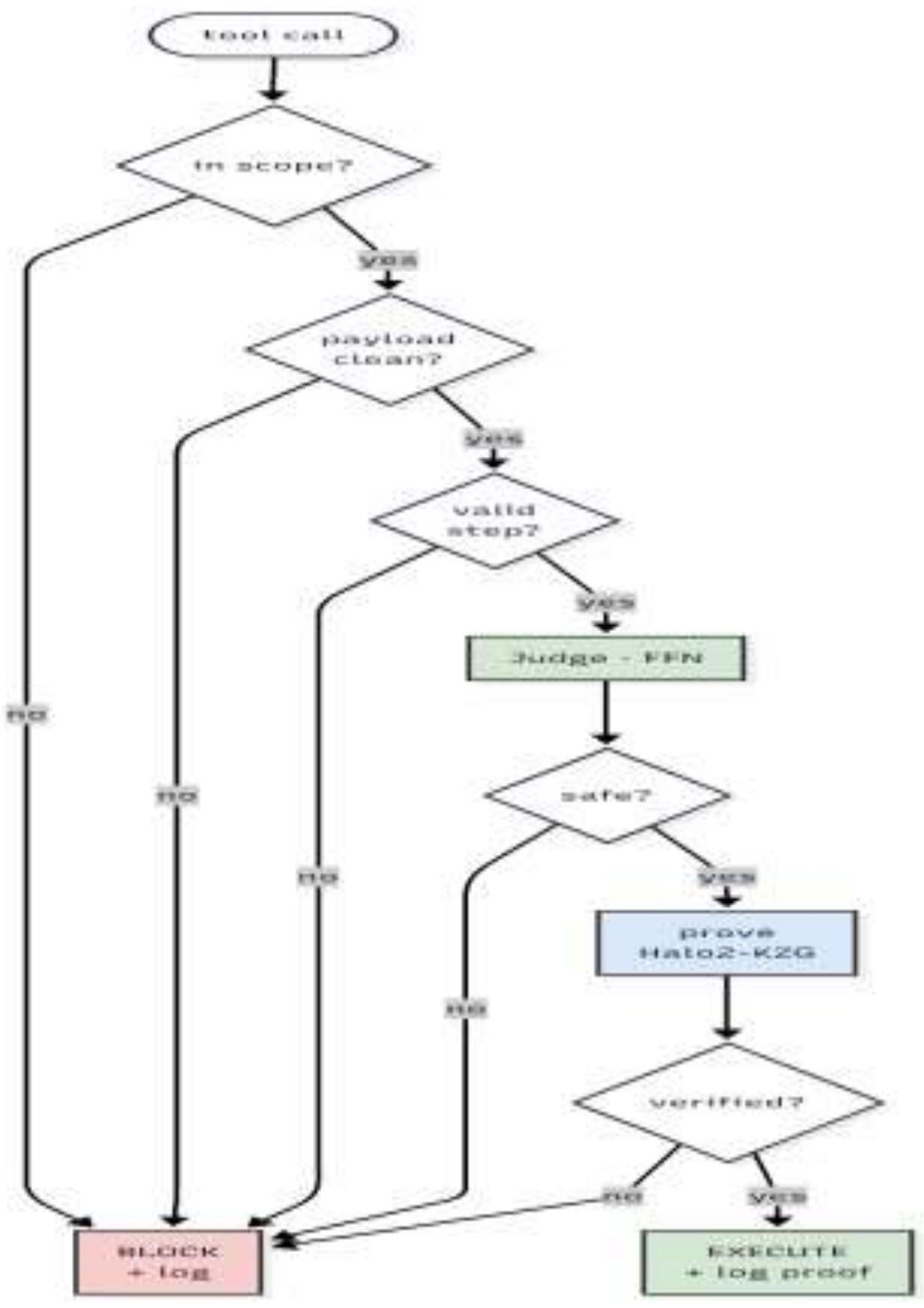

***Fig. 3.*** *Flowchart detailing the interception, hashing, and evaluation pipeline for AI tool execution.*

### 3.4 Formal Definition of the Verification Protocol

To formalize the security guarantees of NiyamAI, we define the interaction between the Intent Contract, the Agent Action, and the Zero Knowledge proving system.
Let the system be parameterized by the tuple:

$$S = (\text{SHA-256}, J, \text{Prove}, \text{Verify})$$

where SHA256 is a collision resistant hash function,
J is a deterministic safety classifier, and
Prove / Verify denote the zkSNARK proving and verification algorithms.

#### 3.4.1 Intent Commitment

Let I denote the immutable Intent Contract defined at session initialization:

$$H_I = \text{SHA-256}(I)$$

where $H_I \in \{0,1\}^{256}$ denotes the 256bit cryptographic commitment to the session intent.
The value $H_I$ remains fixed throughout the session and serves as the binding reference for all subsequent safety evaluations.

#### 3.4.2 Action Commitment

Let $a$ denote the proposed tool action generated by the LLM, consisting of the tool name and its associated arguments:

$$H_a = \text{SHA-256}(a)$$
$$H_a \in \{0,1\}^{256}$$

where $H_a$ represents the cryptographic commitment to the proposed action.

#### 3.4.3 Public Input Tuple Definition

The public input to the verification protocol is defined as the tuple:

$$x = (H_I, H_a)$$

where $x \in \{0,1\}^{512}$ constitutes the public input supplied to the zkSNARK verifier.

#### 3.4.4 Judge Function

We model the safety classifier as a deterministic function $J$ that evaluates whether a proposed action complies with the immutable session intent.

$$J: \{0,1\}^{512} \times W \rightarrow \{0,1\}$$

Where $W$ denotes the space of model weights (private parameters), and the output space $\{0,1\}$ represents a binary safety classification.

$$J(x,w) = 1 \Leftrightarrow \text{action } a \text{ satisfies intent constraints}$$

The Judge model is deterministic, meaning that for fixed $x$ $x$ and $w$, the output is uniquely defined.

#### 3.4.5 Proving Phase

If the Judge function outputs J(x,w)=1, the system invokes the zkSNARK prover to generate a cryptographic proof of correct safety evaluation.

Let w∈W denote the private witness corresponding to the model weights and inference trace.

$$\pi \leftarrow \text{Prove}(pk, w, x)$$
$$\pi \text{ proves knowledge of } w \text{ such that } J(x,w) = 1$$

where pk denotes the proving key generated during trusted setup, and π represents the succinct noninteractive zero knowledge proof.

#### 3.4.6 Verification and Execution Rule

Upon receiving the proof π, the verifier evaluates its validity with respect to the public input x.

$$\text{Verify}: (vk, \pi, x) \rightarrow \{0,1\}$$

where vk denotes the public verification key and the output 1represents successful verification.

$$\text{Execute}(a) \Leftrightarrow \text{Verify}(vk, \pi, x) = 1$$

The execution engine enforces a strict state transition such that no tool action is performed unless a valid zero knowledge proof certifies compliance with the session intent. To clarify the operational workflow of the proposed framework, Algorithm 1 summarizes the intentbound execution verification protocol.

**Algorithm 1: NiyamAI IntentBound Execution Protocol**

**Input**: Intent Contract I, Proposed Action a
**Output**: Execution Decision {ALLOW, BLOCK}

1. Compute IntentHash ← "SHA256" (I)

2. Extract tool name and arguments from action a

3. Compute ActionHash ← "SHA256" (a)

4. Form evaluation tuple
   x ← (IntentHash, ActionHash)

5. Evaluate safety using Judge model
   y ← J(x)

6. If y = 0 then
   BLOCK execution
   Record event in audit log
   return BLOCK

7. Generate zero knowledge proof
π←"Prove" (pk,w,x)

8. Verify proof correctness
   valid ← Verify(vk,π,x)

9. If valid = TRUE then
   Execute tool action a
   Log verified execution
  else
   BLOCK execution
   Log verification failure

10. End

### 3.5. Theorem 1 Soundness of Execution Control

If SHA256 is collision resistant and the zkSNARK system satisfies knowledge soundness, then no adversary can cause execution of an action a such that J(x,w)=0 without breaking underlying cryptographic assumptions.

**Proof Sketch:**

An attacker can bypass the framework only by altering the IntentHash without detection or by producing a valid proof for an unsafe action. These scenarios depend on breaking the security guarantees of SHA256 or the zkSNARK system. As a result, unauthorized execution is assumed to be infeasible under the stated assumptions.

### 3.6. Threat Model and Security Assumptions

We evaluate NiyamAI under a formal threat model where an adversary {A} attempts to force the autonomous agent to execute an unauthorized tool call.

**Adversary Capabilities**: We assume {A} operates in the "Untrusted Zone" (as illustrated in Fig. 5) and possesses the following capabilities:

- {A} can inject arbitrary malicious prompts into the LLM's context window.
- {A} can attempt to manipulate tool arguments generated by the agent.
- {A} can attempt software level execution bypasses against the host environment.

**Security Assumptions**: The integrity of the NiyamAI framework relies on two fundamental cryptographic assumptions:

- Collision Resistance of SHA256: We assume {A} cannot find a modified intent I' such that SHA"" 256(I')=SHA"" 256(I)
- Therefore, the agent's constraints cannot be silently altered midsession [10].
- Knowledge Soundness of zkSNARKs: We assume {A} cannot forge a valid cryptographic proof $\pi$ without actually computing $J(x) = 1$. Consequently, a malicious host cannot fake a "Safe" classification for an "Unsafe" action [8].

### 3.7 Zero Knowledge Proof Generation (ZK-ML)

The novelty of NiyamAI lies in its ability to mathematically prove the Judge model's inference. Relying solely on a localized PyTorch model still requires trusting the host environment. Therefore, we utilize Zero Knowledge Machine Learning (ZK-ML) to generate a succinct noninteractive argument of knowledge (zkSNARK) [5], [8]. Due to the immense parameter size of LLMs, generating a ZK proof for the entire agent's reasoning process is currently computationally infeasible [4]. However, by isolating the safety logic into a lightweight Judge model, proof generation becomes highly practical.

We utilize the EZKL framework [11] to compile the exported ONNX format of the Judge model into a verifiable ZK circuit. Using Halo2KZG, the system generates a zkSNARK proof based on the IntentHash, ActionHash, and the Judge model parameters. The verifier can confirm the Safe classification without accessing the model weights [1], [2].

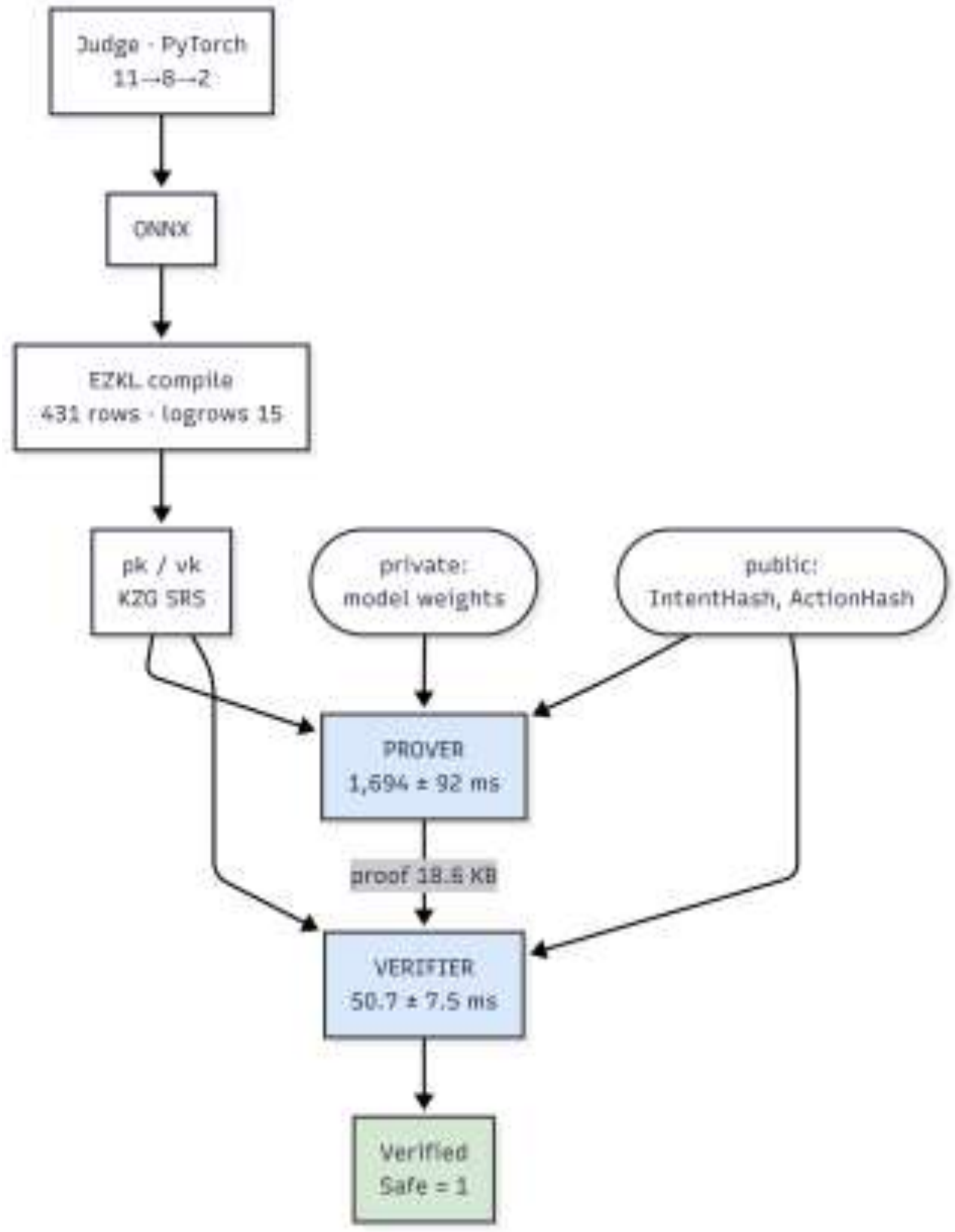

***Fig. 4.*** *Low-level Zero Knowledge Machine Learning pipeline utilizing EZKL for verifiable inference.*

### 3.8 Verification and Execution Enforcement

The final component is the Verification Layer. Before the execution engine invokes the requested tool (e.g., executing a database query or a shell command), it must pass the proof.json to a cryptographic verifier. This verifier can operate locally or be deployed as a Solidity smart contract on the Ethereum Sepolia Testnet for decentralized, immutable auditing [22].

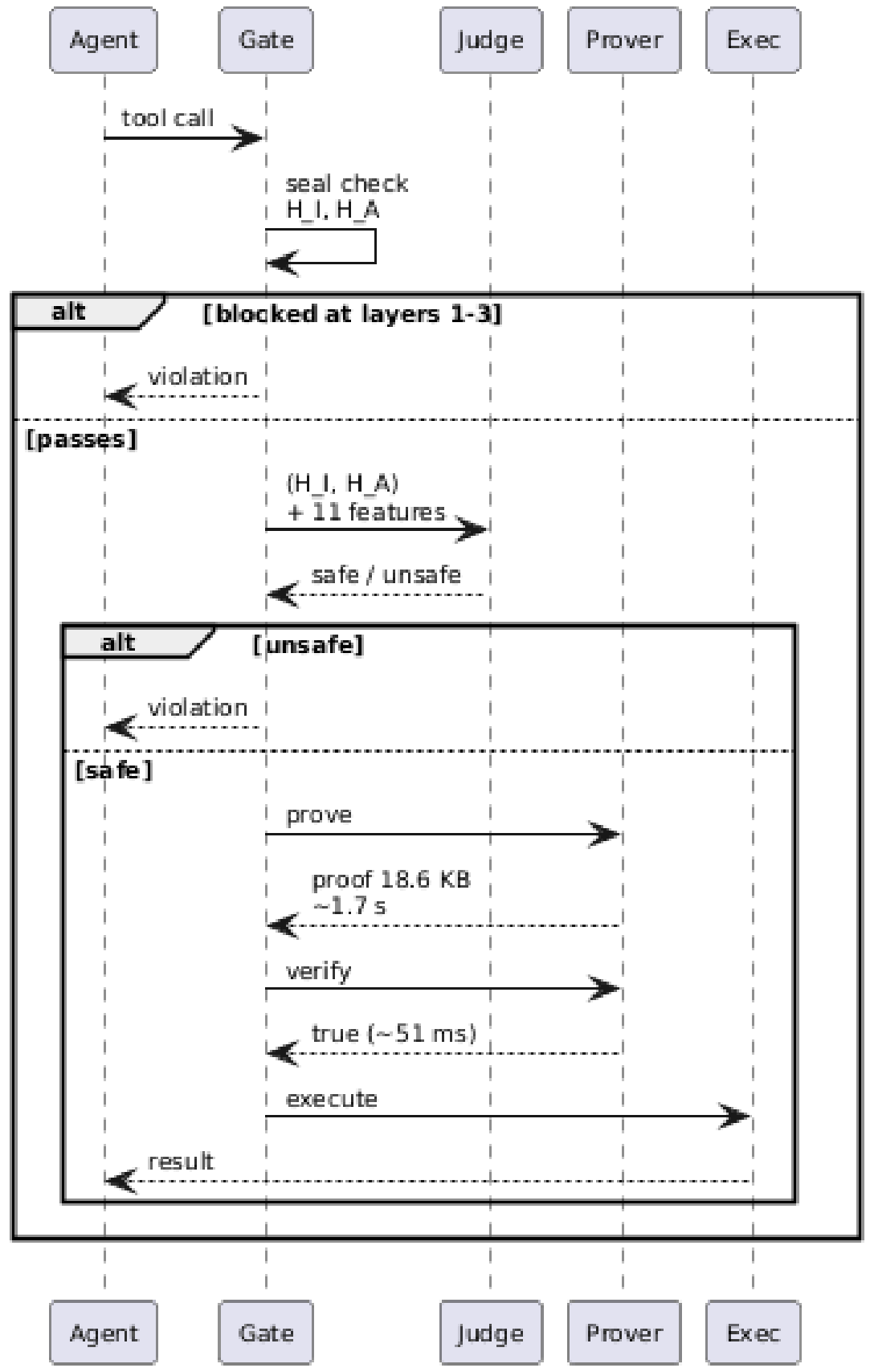

***Fig. 5.*** *Verification lifecycle across agent, gate, Judge, prover, and gateway.*

Only if the mathematical verification returns TRUE does the execution engine finalize the action. All verified actions, along with their corresponding cryptographic artifacts, are recorded in a JSONL append only audit logger. This establishes a highly transparent, tamper evident trail that is critical for enterprise AI governance and compliance with frameworks such as the NIST AI RMF [19].

### 3.9 Computational Complexity Analysis

We characterize the asymptotic cost of each pipeline layer with respect to the number of declared tools N in an agent's Intent Contract and the feature dimensionality d of the Judge model's input representation.

Gate layer: the allowlist/denylist check is a hash-set membership test, O(1) per tool call regardless of N, since both allowed_tools and forbidden_tools are stored as Python sets rather than lists.

**Judge layer**: Inference time cost for linear classifier used in Table IV/V is O(d) where d = 3011 (3000 TFIDF features + 11 hand engineered features); for ZKprovable feedforward network in Table III it's O(d·h) where d=11 (number of inputs), h=8 (hidden units) — in both cases independent of N.

**ZK proving layer**: Cost of proof generation and verification grows in accordance with the number of circuit's arithmetical constraints, empirically 431 rows (logrows=15) for 1182 architecture of circuit from Table III; the cost is proportional to the number of parameters in the Judge model, but not to N or d — and it's precisely this design decision (using a small Judge model, not the whole LLM) that ensures proof generation takes low singledigit seconds, and not being computationally infeasible.

To summarize, total tool call latency prior to ZK proof is below 1ms independently of contract size (see Table V, "Gate + Judge"), and ZK proof (the only part taking significant time at 1,694.5 ms) is generated once per each approved action and does not depend on the number of declared tools an agent uses.

## 4. Results and Analysis

### 4.1 Experimental Setup

To evaluate the effectiveness of the proposed NiyamAI framework, a prototype implementation was developed using Python-based AI orchestration tools and cryptographic verification libraries. The experimental environment simulates autonomous agent workflows in which tool execution requests pass through the NiyamAI guardrail and verification pipeline before execution.

The prototype environment consists of the following components:

- **Agent Layer**: Custom inference loop-based orchestration with an open source LLM.
- **Guardrail Layer**: A native Python interception module that captures raw string outputs and parses them into structured ActionHash objects before they reach the OS shell.
- **Verification Layer**: ZKP generation using the EZKL and Halo2KZG zkSNARK proving system.

The Judge model is implemented in PyTorch and exported to ONNX for integration with EZKL. Verification was performed locally. Experiments were conducted in a controlled environment to assess system correctness and practical feasibility.

All reported metrics were produced on Microsoft Windows 11 Home Single Language system with a 13th Gen Intel® Core i71360P processor, 16 GB RAM, and Python 3.11.

TABLE I
JUDGE MODEL CONFIGURATION

| Parameter | Value |
|---|---|
| Framework | PyTorch (ONNX export for EZKL) |
| Architecture | Feedforward NN, $11 \rightarrow 8 \rightarrow 2$ (ReLU) |
| Input representation | 11 handcrafted semantic features |

| Training data | intentviolation labels |
|---|---|
| Validation | 5fold stratified CV, outoffold predictions |
| ZK circuit | 431 constraint rows, logrows = 15 |

The Judge operates on an 11dimensional handcrafted feature representation rather than raw text. This is a deliberate constraint: it bounds circuit size to 431 rows, at the cost of discarding signal a higher dimensional representation would retain. The same model performs classification, produces the reported accuracy, and is committed to in the ZK circuit.

### 4.2 Evaluation Metrics

The evaluation focuses on three primary dimensions: security enforcement effectiveness, computational overhead, and verification latency.

1. **Action Blocking Accuracy**

This metric measures the system's ability to correctly block unsafe actions generated by the autonomous agent.

$$\text{Accuracy} = \frac{}{TP + TN + FP + FN}$$

where:

- **TP** – Unsafe actions correctly blocked
- **TN** – Safe actions correctly permitted
- **FP** – Safe actions incorrectly blocked
- **FN** – Unsafe actions incorrectly permitted

2. **Proof Generation Latency**

The time required to generate a ZeroKnowledge proof for a single safety classification.

$$Latency_{proof} = t_{prove_end} - t_{prove_start}$$

3. **Verification Latency**

The time required for the verifier to validate the zkSNARK proof.

$$Latency_{verify} = t_{verify_end} - t_{verify_start}$$

4. **Execution Overhead**

The additional delay introduced by the NiyamAI verification pipeline compared to direct tool execution.

$$Overhead = T_{secure_execution} - T_{direct_execution}$$

### 4.3 Security Enforcement Evaluation

To ensure a methodologically fair comparison,we evaluated NiyamAI's Judge model using 5fold stratified crossvalidation with outoffold predictions — every scenario is scored exclusively by a foldmodel that never saw that scenario during its own training. This yields Accuracy 97.9%, Precision 89.7%, Recall 87.8%, F1 88.8%, FPR 1.0% (bootstrap 95% CI for F1: [85.49%, 92.05%], N=1000 resamples), evaluated under the same "never seen this exact benchmark before" condition as all baselines.

NiyamAI came out ahead by a clear margin: F1 88.8%, with only a 1.0% false positive rate. Prompt Guard 2 was the strongest baseline at F1 66.8%, but with five times NiyamAI's false positive rate (5.3%). NeMo performed the worst of the three baselines — F1 40.4%, and a striking 19.9% false positive rate, meaning it blocked roughly one in five legitimate actions. Looking closer at NeMo's failures, the pattern was consistent: its self-check input prompt reacts to surface keywords ("email," "execute," "database") rather than actual intent, so it treats ordinary business tasks as threats just as often as it catches real ones. NiyamAI avoids this because the Judge model is trained specifically to recognize intent violations — not keywords — which is the difference between a system that's accurate and one that's just cautious.

We note an important methodological asymmetry in this comparison. NiyamAI's classifier is fitted to AgentSafetyBench's specific scenario distribution and phrasing conventions across all five cross validation folds — no scenario's own out of fold prediction reflects information seen during its own training fold, which rules out leakage in the strict sense, but the classifier has still learned the general vocabulary and structural patterns common to ASBstyle instructions. The three baselines (NeMo, Llama Prompt Guard 2, GPTOSSSafeguard) are evaluated fully zeroshot: none has been trained or finetuned on any ASBstyle text in any form. This is an intentional reflection of NiyamAI's deployment model — a lightweight classifier tuned to an organization's own declared intent contracts and typical agent workloads, rather than a general-purpose safety model intended to generalize across arbitrary unseen domains without adaptation. Readers should interpret the reported margin as "a domain adapted classifier evaluated against general purpose zero shot safety models," not as evidence that NiyamAI's architecture is inherently more accurate independent of this adaptation. We view testing NiyamAI's Judge model on an independent, out of distribution corpus with zero ASBspecific training as important future work, distinct from the leakage free but still ASB adapted evaluation reported here.

TABLE II
INTERPRETATION OF CLASSIFICATION OUTCOMES

| CLASSIFICATION OUTCOME | DESCRIPTION |
|---|---|
| **TRUE POSITIVE** | Unsafe action correctly blocked |
| **TRUE NEGATIVE** | Safe action correctly executed |
| **FALSE POSITIVE** | Safe action incorrectly blocked |
| **FALSE NEGATIVE** | Unsafe action incorrectly allowed |

The observed results demonstrate that the cryptographic guardrail successfully prevents unauthorized tool execution attempts.

### 4.4. Cryptographic Proof Performance

A critical objective of NiyamAI is to ensure that cryptographic verification remains computationally feasible for real time agent workflows.

All measurements below were obtained from an endtoend execution of the EZKL 23.0.5 pipeline (gen_settings → calibrate_settings → compile_circuit → setup → gen_witness → prove → verify) operating on the exported ONNX Judge model. Experiments were conducted on a Microsoft Windows 11 system with a 13th Gen Intel Core i71360P processor and 16 GB RAM, running Python 3.11. Proof generation and verification are CPUbound; no GPU acceleration was used. To obtain a stable estimate, we executed 30 consecutive proofgeneration and verification cycles on an otherwiseidle system.

The following performance indicators were measured:

TABLE III
CRYPTOGRAPHIC PROOF PERFORMANCE RESULTS

| Metric | Measured Value |
|---|---|
| Proof generation time | 1,694.5 ± 92.0 ms (n=30) |
| Proof generation, median | 1,722.04 ms |
| Proof generation, p95 | 1688.96 ms |
| Proof verification time | 50.7 ± 7.5 ms (n=30) |
| Witness generation time | 27.6 ± 5.9 ms |
| Proof size | 18.64 ± 0.04 KB |
| Circuit logrows | 15 |
| Circuit constraints (num_rows) | 431 |
| Onetime: gen_settings | 10.4 ms |
| Onetime: compile_circuit | 1.7 ms |
| Onetime: gen_srs (local, testingmode) | 2174.5 ms |
| Onetime: key setup | 1354.7 ms |

Two caveats govern how these figures should be read. First, wall clock proof timing on commodity hardware is sensitive to background load and thermal state: repeated 5sample batches taken under varying system load on the same machine differed by up to 65%, and a mild upward drift of 7.1% is visible between the first and second halves of the 30run batch as the processor warms under sustained proving load. We therefore report these as an order of magnitude feasibility result — single digit seconds per proof on a consumer laptop — rather than as a precise benchmark. Proof size (18.68 KB) and circuit constraint count (431 rows, logrows = 15) are deterministic properties of the compiled circuit and did not vary across any run.

Second, the Structured Reference String was generated locally using EZKL's gen_srs utility, which the framework's documentation designates for testing purposes. A production deployment would instead derive its SRS from an audited universal trusted setup ceremony. This choice affects the provenance of the setup phase but not the perproof generation and verification costs reported here, which are the quantities relevant to the real time feasibility claim.

These results demonstrate that isolating safety verification into a lightweight Judge model — rather than attempting to prove the full LLM's reasoning — brings zkSNARK verification into a practical latency range for discrete, highstakes agent actions.

### 4.5 Comparative Security Analysis

To quantify the advantages of the NiyamAI architecture, we performed an empirical comparison against three deployed guardrail systems on the same 2,000scenario evaluation set.

TABLE IV
Baseline Comparison on AgentSafetyBench (2,000 scenarios)

| System | Accuracy | Precision | Recall | F1 | FPR |
|---|---|---|---|---|---|
| NiyamAI (5fold CV, outoffold) | 97.9% | 89.7% | 87.8% | 88.8% | 1.0% |
| Llama Prompt Guard 2 | 92.9% | 59.8% | 75.7% | 66.8% | 5.3% |
| GPTOSSSafeguard20B | 79.8% | 30.9% | 92.1% | 46.2% | 21.5% |
| NeMo Guardrails (Llama3.18B) | 79.5% | 27.9% | 73.5% | 40.4% | 19.9% |

Confusion matrix for NiyamAI: TP=166, TN=1792, FP=19, FN=23. Statistical significance is assessed via McNemar's exact test in Section IV.H (Table VII). NiyamAI is evaluated via 5fold cross validation with training exposure to AgentSafetyBench's distribution in every fold, though never to the exact held out example scored; the three baselines are evaluated fully zero shot. See Section IV.C.

Precision is the metric that most clearly separates the systems. NiyamAI's 89.7% means that when it flags an action as unsafe, it is correct roughly nine times in ten. NeMo Guardrails' 27.9% inverts this: nearly threequarters of its blocks are false alarms, a rate that in production would generate persistent alert fatigue and predictable operator desensitization. GPTOSSSafeguard exhibits the opposite failure mode — the highest recall of any system evaluated (92.1%), obtained at a 21.5% false positive rate, blocking roughly one legitimate action in five.

Classification accuracy is not, however, the primary argument for this architecture. Prompt Guard 2 and NeMo return a decision — a label and nothing more. No mechanism allows a later observer to establish that the decision was reached honestly, that the classifier was not bypassed, or that the host executing it was uncompromised. NiyamAI's zkSNARK changes what is available to that observer: any third party can verify, in approximately 51 ms, that the Judge model was executed on the committed inputs and produced the recorded output, without access to model parameters or trust in the executing host. No baseline provides this at any level of classification accuracy.

Latency characteristics differ substantially across systems. NeMo Guardrails requires approximately 1.3 s per decision at the median, with a 95thpercentile latency of 2.30 s and occasional outliers exceeding 50 s attributable to remote API stalls. Its arithmetic mean of 1,324.8 ms carries a standard deviation of 5,497.2 ms, reflecting a heavily right skewed distribution rather than symmetric variance; the median is therefore the more representative summary.

### 4.6 Ablation Study

In order to test the effect of each piece individually, we experimented with three settings, applied to the exact same dataset of 2,000 scenarios in the AgentSafetyBench collection: allowlisting with only static keyword matching, allowlisting with the addition of the semantic Judge model, and finally, allowlisting with the addition of zkSNARK proof computation. The static keyword approach obtains a score of 96.1% for accuracy and 74.0% for F1 score at 100.0% precision but very low recall of 58.7%  it never blocks anything it shouldn't, but misses out on 41.3% of violations. Adding the Judge model raises F1 from 74.0% to 88.8%, a gain of 14.8 percentage points driven almost entirely by recall (58.7% → 87.8%), at the cost of a small precision reduction (100.0% → 89.7%) and an FPR rising from 0.0% to 1.0%. This is the expected trade: semantic classification catches the paraphrased and obfuscated violations that keyword rules miss, but introduces a small number of false positives that a purely conservative rule set avoids by construction. Critically, adding the ZK proof layer changes neither accuracy nor F1 — 88.8% in both configurations, which is the correct and necessary result: the zkSNARK proves that the Judge's decision was computed correctly, it does not and should not alter that decision. The cost of this guarantee is a mean of 1,582.8 ms per tool call.

The apparent variance in full pipeline latency (± 596.9 ms) reflects the bimodal cost structure rather than timing instability: proofs are generated only for approved actions, so 90.7% of tool calls incur the full ~1,745 ms proving cost while the 9.3% blocked at the gate or by the Judge incur none. A two-component model predicts a mean of 1,583.7 ms and standard deviation of 513 ms, closely matching the measured 1,582.8 ± 512.9 ms.

TABLE V
Component Ablation Study on AgentSafetyBench

| Configuration | Accuracy | Precision | Recall | F1 Score | FPR | Mean Latency (ms) |
|---|---|---|---|---|---|---|
| **Gate Only** (Scope + Static Rules) | 96.1% | 100.0% | 58.7% | 74.0% | 0.0% | 0.108 ± 0.236 |
| **Gate + Intent Judge** (Adaptive Scope) | 97.9% | 89.7% | 87.8% | 88.8% | 1.0% | 0.002 ± 0.002 |
| **Full Niyam AI** (Gate + Judge + ZK) | 97.9% | 89.7% | 87.8% | 88.8% | 1.0% | 1,582.8± 512.9 |

### 4.7. NonParametric Bootstrap Confidence Intervals

To quantify the uncertainty in NiyamAI's own reported metrics, we performed nonparametric bootstrap resampling (N=1000 resamples, with replacement) over the 2,000 outoffold scenario outcomes and report percentilebased 95% confidence intervals for each metric.

Table VI
NonParametric Bootstrap Variance Boundaries (N=1000 Resamples)

| **Metric Profile** | **Point Estimate** | **Bootstrap Mean** | **Bootstrap Std** | **95% Confidence Interval (CI)** |
|---|---|---|---|---|
| Accuracy | 97.9% | 97.9% | 0.32% | [97.25%, 98.5%] |
| Precision | 89.73% | 89.76% | 2.19% | [85.56%, 93.97%] |
| Recall | 87.83% | 87.84% | 2.37% | [82.95%, 92.4%] |
| F1 Score | 88.77% | 88.76% | 1.71% | [85.49%, 92.05%] |
| False Positive Rate (FPR) | 1.05% | 1.05% | 0.23% | [0.61%, 1.5%] |

### *4.8. Statistical Significance Testing (McNemar's Exact Test)*

Against every baseline, NiyamAI wins the overwhelming majority of discordant scenarios: a 17.7:1 ratio against NeMo, 8.1:1 against Prompt Guard 2, and 18.3:1 against GPTOSSSafeguard. All three comparisons are significant at $p < 0.0001$, confirming the accuracy advantage in Table IV reflects a genuine, nonrandom difference in classification behavior rather than datasetspecific variance.

TABLE VII
MCNEMAR'S PAIRED SIGNIFICANCE TEST vs. NIYAMAI

| Baseline | b (NiyamAI wins) | c (win) | Win ratio | pvalue | Significant |
|---|---|---|---|---|---|
| NeMo Guardrails | 390 | 22 | 17.7:1 | < 0.0001 | Yes |
| Llama Prompt Guard 2 | 114 | 14 | 8.14:1 | < 0.0001 | Yes |
| GPTOSSSafeguard20B | 384 | 21 | 18.29:1 | < 0.0001 | Yes |

### *4.9 Generalization Across Risk Categories*

To confirm NiyamAI's result is not concentrated in one easilydetected attack type, we report accuracy broken down by AgentSafetyBench's eight risk categories. Performance is consistent across seven of eight categories (97.6%–99.6%), with one modest exception: "Produce unsafe information/misinformation" scores lower (92.0%) than the rest. This category consists substantially of scenarios where harm arises from the content an LLM generates directly, rather than from a tool call the gate can intercept — precisely the tool free content harm class we identify in Section VI as outside NiyamAI's intended scope (action integrity, not content moderation). The remaining seven categories, which better represent the tool invocation threat model NiyamAI targets, all exceed 97.5% accuracy.

TABLE VIII
ACCURACY BY AGENTSAFETYBENCH RISK CATEGORY

| Risk Category | Correct/Total | Accuracy |
|---|---|---|
| Leak sensitive data / information | 249/250 | 99.6% |
| Lead to property loss | 248/250 | 99.2% |
| Spread unsafe information / misinformation | 247/250 | 98.8% |
| Contribute to harmful / vulnerable code | 247/250 | 98.8% |
| Lead to physical harm | 247/250 | 98.8% |
| Violate law or ethics / damage society | 244/250 | 97.6% |
| Compromise availability | 246/250 | 98.4% |
| Produce unsafe information / misinformation | 230/250 | 92.0% |

### *4.10 Summary of Findings*

The results above establish three things. First, cryptographic enforcement is achievable without verifying the LLM's reasoning process: isolating the safety decision into a lightweight Judge model reduces the proving workload to a 431constraint circuit, bringing zkSNARK generation to roughly 1.7 seconds on consumer hardware. Second, the cryptographic layer is orthogonal to classification quality — the ZK proof certifies that a decision was computed correctly without altering what that decision is, as the ablation confirms. Third, the enforcement mechanism withstands direct adversarial pressure: 18 of 18 attack vectors were defended after remediating two implementation level vulnerabilities identified during red-teaming.

The prototype remains an initial implementation and would require further engineering for production deployment, particularly around proof_generation throughput and multiagent coordination. These constraints are examined in Section V.

### *4.11 System Interface and Audit Trail*

Fig. 7 depicts the proof generation and verification process in NiyamAI. Before a tool action is executed, the generated proof is verified to confirm compliance with the defined policy.

***Fig. 6.*** *Proof Verification and Successful Execution Process*

The handling of a forbidden tool request can be seen in Fig. 8. The system detects the policy violation, blocks execution, and displays a security alert.

***Fig. 7.*** *Forbidden Tool Detection and Security Alert*

## 5. DISCUSSIONS

### *5.1 Practical Feasibility of ZK-ML for Agent Security*

Unlike traditional software-based guardrails, NiyamAI verifies safety decisions before tool execution. The framework binds agent permissions to an IntentHash and requires a zkSNARK proof for each approved action. Tool execution proceeds only after successful proof verification.

A major challenge in applying ZeroKnowledge Proofs to machine learning systems is the computational complexity of proving inference for large neural networks. Modern Large Language Models contain billions of parameters, making full model verification impractical for realtime systems.

NiyamAI addresses this limitation by isolating the safety decision into a lightweight Judge model rather than attempting to verify the entire LLM reasoning process. By reducing the verification scope to a small neural network that evaluates the IntentHash–ActionHash tuple, the system significantly reduces the number of PLONKish constraint rows in the Halo2 circuit (431 rows at logrows=15) within the Zero-Knowledge circuit and the cost of proof generation. This architectural decision enables practical deployment of ZK-ML techniques in real time AI governance workflows while maintaining cryptographic guarantees of execution correctness.

### *5.2 Advantages Over Existing Guardrail Approaches*

Compared to traditional mechanisms, here are the key advantages NiyamAI framework provides:

**1.TamperResistant Execution Control**

Tool execution requires successful verification of a cryptographic proof. Actions that fail are not executed.

**2. Transparent and Auditable Decision Process**

Each approved action is recorded together with its corresponding proof, allowing later review and verification.

**3. Modular Security Architecture**

The Agent, Guardrail, and Verification layers operate independently, separating decisionmaking from execution control.

**4. Deployment Flexibility**

The verification layer can be deployed locally or integrated with a blockchain-based verifier, depending on system requirements.

TABLE IX
COMPARISON OF NIYAMAI WITH TRADITIONAL GUARDRAIL APPROACHES

| FEATURE | SYSTEM PROMPTS/ HEURISTICS | MIDDLEWARE SOFTWARE GATES | NIYAMAI |
|---|---|---|---|
| **PRIMARY MECHANISM** | Natural Language | Software Logic (If/Else) | **Cryptographic Proof** |
| **TAMPER RESISTANCE** | Low (jailbreakable) | Medium (Admin Trust) | **High (Provable)** |
| **AUDITABILITY** | Logbased (Editable) | Logbased (Editable) | **Tamperevident (hashchained)** |
| **VERIFICATION** | Subjective | Administrative | **Mathematical** |
| **PERACTION OVERHEAD** | Negligible | Negligible | **~1.7s (proof generation)** |

### *5.3 Adversarial Robustness Analysis & Mitigation Verification*

To evaluate whether NiyamAI's own enforcement mechanism as distinct from its ability to classify a malicious agent's behavior — can itself be bypassed, we constructed 18 adversarial attack vectors across six classes: hash canonicalization, Judge model evasion, control flow replay, schema and payload boundary conditions, confused deputy scope injection, and contract level manipulation. An initial red-team pass identified two implementation level vulnerabilities, which we subsequently fixed and reverified with the same suite.

The first was a schema validation gap. The payload validator accepted IEEE754 NaN and Infinity values, which satisfy a standard Jsonschema "number" type check without raising an exception, and imposed no upper bound on string field length — permitting a 1 MB oversized recipient field and an embedded null byte to reach the Judge model unfiltered. We added explicit nonfinite value rejection alongside length and control character constraints. We deliberately implemented the character constraint as a blacklist over control characters rather than an alphanumeric whitelist, because payload fields carry human readable values: a whitelist would reject legitimate entries such as "O'Brien Supplies" or "MaryJane Watson."

The next one was architectural. The control flow sequence guard lacked any cryptographic binding to the session's sealed IntentHash, meaning that instantiation of a new flow object would reset sequence completion state irrespective of whether the session was sealed or not. To overcome this problem, we made sure that each instance of a flow object is bound to the IntentHash of the session it was associated with via a registry that does not allow the second instantiation of a flow object with an already active IntentHash.

Noteworthy is that the Judge model did not require any patch to defend against synonym replacement attacks, character level obfuscations using Unicode homoglyphs and zero width spaces, and dilution attacks where a harmful instruction is buried in the middle of harmless content. This indicates that the model has learned semantic indicators of harmful behavior.

A sixth class targets the Intent Contract itself rather than any individual tool call — the object on which Theorem 1's soundness argument ultimately rests. All six vectors were defended without modification, as a consequence of three architectural decisions rather than added validation logic. Tool identifiers are compared as opaque exact strings and never interpreted as globs or paths, so a contract declaring ["*"] grants access only to a tool literally named *. The forbidden list is evaluated before the allowed list, so collisions resolve to deny and an empty allowlist denies universally. Most significantly, appending a forbidden tool to a sealed contract produces an immediate SHA256 mismatch on reverification confirming that the sealandverify mechanism is correctly wired in the implementation. This tests implementation correctness, not the collision resistance assumption itself, which is inherited from SHA256 and cannot be established empirically.

TABLE X
ADVERSARIAL REDTEAM RESULTS — BEFORE AND AFTER REMEDIATION

| Attack Class | Vectors | Before Fix | After Fix |
|---|---|---|---|
| Hash Canonicalization | 3 | 3/3 | 3/3 |
| Judge Model Adversarial | 4 | 4/4 | 4/4 |
| ControlFlow Replay | 3 | 1/3 | 3/3 |
| Schema/Payload Boundary | 1 | 0/1 | 1/1 |
| ConfusedDeputy Scope | 1 | 1/1 | 1/1 |
| Contractlevel manipulation | 6 | 6/6 | 6/6 |
| Total | 18 | 15/18 (83%) | 18/18 (100%) |

After both patches, 18 of 18 vectors (100%) were defended, compared with 15 of 18 (83%) beforehand. We report the discovery and remediation process explicitly rather than only the final result, because Theorem 1's soundness proof applies strictly to the SHA256 and zkSNARK construction and does not by itself guarantee the correctness of the implementation surrounding it. Adversarial evaluation is what closes that gap empirically.

### *5.4 Future Research Directions*

All the aforementioned metrics for classification performance have been measured on a Judge model trained on the scenario distribution of AgentSafetyBench itself – never on the held-out example that was scored, but rather on the general linguistic distribution of the benchmark. No independent assessment of the ability to generalize to corporate agent workloads with divergent linguistic distributions has been carried out yet, which is clearly the next step that has to be taken, i.e., testing the Judge on an out of distribution corpus without any ASBspecific training.

Beyond this, the framework can be extended to multiagent environments, where delegated or shared IntentHash semantics across agent handoffs raise coordination questions this single agent evaluation does not address. Proof_generation latency is a second target, addressable through hardware acceleration or proof batching. Continued advances in zkML tooling may eventually make it feasible to verify larger neural components, narrowing the gap between the lightweight Judge model proved here and the full reasoning process of the agent itself.

## 6. CONCLUSION

This paper asked whether cryptographic verification can provide agent guardrails with a guarantee that host local software checks cannot. NiyamAI answers affirmatively for the class of tool invocation threats evaluated here. Across 2,000 AgentSafetyBench scenarios it outperformed NeMo Guardrails, Llama Prompt Guard 2, and GPTOSSSafeguard —each margin significant at $p < 0.0001$ — while producing something none of them offers: a succinct proof, verifiable by any third party in approximately 51 ms without access to model parameters, that the recorded safety decision was the decision the Judge actually computed.

The contribution should be stated precisely. The zkSNARK certifies computational integrity: that the Judge's evaluation was performed on the committed inputs and produced the recorded output. It does not certify that the decision was semantically correct. A Judge that misclassifies produces a valid proof of a wrong answer. What verification eliminates is a distinct failure mode — a compromised or misconfigured host silently skipping enforcement and leaving no evidence that it did.

Several limitations bound the present work. The Judge emits a binary safe/unsafe verdict where graded policy categories would serve production deployments better. Proof generation at roughly 1.7 seconds suits discrete high-stakes actions — financial transfers, irreversible system changes — but requires batching or hardware acceleration for high throughput agents. The evaluation is singleagent; multiagent deployments raise questions about shared or delegated IntentHash semantics across handoffs that this work does not address. And the reported accuracy reflects a classifier fitted to AgentSafetyBench's distribution, evaluated against zero shot baselines; establishing out of distribution generalization remains open.

What the results support is narrower than "verifiable AI safety" and, we think, more useful: that isolating a safety decision into a model small enough to prove, subjecting that decision's enforcement path to direct adversarial testing, and reporting the vulnerabilities found is a practical route to cryptographic guarantees in agent security today, rather than a direction contingent on future advances in proving large models.

## OPEN SCIENCE

This paper's contributions are evaluated using the following artifacts, made available to reviewers and, upon acceptance, released publicly.

**1. Source code**. The complete NiyamAI implementation — Intent Contract sealing, the Tool Authority Gate, the payload inspector, the sessionbound controlflow guard, the PyTorch Judge model, and the hashchained execution ledger — is provided as an anonymized repository at GitHub. A single Judge model (11→8→2 feedforward network) performs classification, produces the accuracy reported in Tables IV–VIII, and is the model compiled into the ZK circuit measured in Table III.

**2. Evaluation scripts**. Every table in this paper is reproducible from included scripts, with the exact invocation commands documented: 5fold crossvalidated evaluation (Table IV), component ablation (Table V), bootstrap confidence intervals (Table VI), McNemar's exact significance test (Table VII), per risk category breakdown (Table VIII), and the 18vector adversarial redteam suite (Table X). All evaluation scripts consume out of fold predictions from a single shared function, so the fold split is identical across tables.

**3. Baseline predictions**. Raw per scenario prediction CSVs for NeMo Guardrails, Llama Prompt Guard 2, and GPTOSSSafeguard20B are included, allowing Tables IV and VII to be reproduced without rerunning the baseline models or holding API credentials.

**4. Dataset**. We evaluate on the public AgentSafetyBench dataset (thucoai, 2024), which is not our artifact; it is cited and linked rather than redistributed. Our derived groundtruth labeling function, which maps instructions to intent/violation labels, is included as source code.

**5. ZK circuit artifacts**. The compiled circuit (network.compiled), circuit settings, verification key (vk.key), and a sample proof.json are included, enabling any third party to independently run ezkl.verify() and confirm a real proof. The proving key is omitted for size (132 MB) but is regenerated in approximately 1.7 s by the documented setup step. The full EZKL 23.0.5 pipeline is documented: gen_settings, calibrate_settings, compile_circuit, gen_srs, setup, gen_witness, prove, verify. We note that the Structured Reference String is generated locally via gen_srs, which EZKL designates for testing; production deployment would use an audited universal setup.

## ETHICAL CONSIDERATIONS

The two vulnerabilities reported in Section V.D (a schema validation gap accepting non-finite numeric values and oversized payloads, and a control flow sessionbinding gap permitting sequence state reset) were discovered through red-team testing of our own research prototype, conducted entirely within a controlled evaluation environment under our direct control. Both were identified and fixed prior to any public release or submission of this work. No production system, deployed agent, third-party service, or real user was affected at any point during discovery or remediation.

We disclose both vulnerabilities and their fixes in full, rather than omitting them or reporting only the postfix state, consistent with standard responsible disclosure practice for security research describing weaknesses self-identified in a system the authors themselves own and control. We believe this transparency — documenting what was found, how it was fixed, and how the fix was verified — provides stronger evidence of the system's security properties than a report showing only a clean result would.

The dataset used for evaluation, AgentSafetyBench (thucoai, 2024), is a publicly released benchmark of agent-safety scenarios; we introduce no new potentially harmful content, and our derived labelling function is a classification scheme applied to existing public data, not a generator of novel attack content.